\newcommand{\upcite}[1]{\textsuperscript{\textsuperscript{\cite{#1}}}}
\RequirePackage{fix-cm}
\documentclass[smallcondensed]{svjour3}     
\smartqed  
\usepackage{graphicx}
\usepackage[numbers,square,sort]{natbib}
\usepackage[justification=centering]{caption}
\begin{document}

\title{Patch Network for medical image Segmentation
}


\author{Weihu Song          \and
        Heng Yu             \and
        Jianhua Wu
}


\institute{Beihang University\at
              \email{weihusong@buaa.edu.cn}   \at
              Beijing, 100191, China
          \and
          Carnegie Mellon University \at
            \and
            Nankai University\at
}

\date{Received: date / Accepted: date}

\maketitle

\begin{abstract}
Accurate and fast segmentation of medical images is clinically essential, yet current research methods include convolutional neural networks with fast inference speed but difficulty in learning image contextual features, and transformer with good performance but high hardware requirements. In this paper, we present a Patch Network (PNet) that incorporates the Swin Transformer notion into a convolutional neural network, allowing it to gather richer contextual information while achieving the balance of speed and accuracy. We test our PNet on Polyp(CVC-ClinicDB and ETIS- LaribPolypDB), Skin(ISIC-2018 Skin lesion segmentation challenge dataset) segmentation datasets. Our PNet achieves SOTA performance in both speed and accuracy.
\end{abstract}

\section{Introduction}\label{sec1}
Colorectal cancer (CRC) is the world's third most frequent cancer after lung cancer, and the majority of CRC cases are caused by polyp transformation. It can successfully prevent polyp from changing into CRC if it can be detected and removed in time at an early stage, Therefore, the precise prediction of the lesion area becomes a central task. The most prevalent treatment technique is colonoscopy, which is often performed manually by trained clinicians. Because finding polyps is difficult and time-consuming, automated precision segmentation approaches are critical. Skin lesions are another common and under-appreciated ailment. While most Skin lesions are just mildly detrimental to the body, owing to their diverse kinds, some Skin lesions can become permanent losses if not treated promptly and lead to other diseases and even progeny. The exact segmentation of a Skin lesion's site, which offers clinicians the location information about diseased areas for their follow-up work, is the emphasis of Skin lesion treatment. In summary, CRC and Skin lesions both require accurate segmentation of the lesion location. Most of the current research methods effectively prove the feasibility of deep learning for medical image segmentation. In medicine, this segmentation field's accuracy and speed requirements are rigorous, and these methods are challenging to meet the needs. Suitable performance methods will be relatively high hardware requirements that consume a lot of time, and detection speed methods will be some of the target object details features ignored. These problems in the medical field need to be effectively addressed. To address this, we propose a Patch block to incorporate the idea of Swin Transformer, which has good performance, into the fast convolutional neural network(CNN) for segmentation to obtain more contextual information in the form of patching, effectively merging the advantages of the two and at the same time, we propose a novel lightweight network Patch Network for medical image segmentation, we test our model on three datasets, intersection over union(IOU) and Dice similarity coefficient(Dice) reach 0.9332 and 0.9599 on CVC, IOU and Dice reach 0.9405 and 0.9646 on ETIS, IOU and Dice get 0.8946 and 0.9340 on Skin, and The number of model parameters and floating point operations per second(FLOPs) is only 1/10 of that of UNet++, while the fps is more than three times that of UNet++.

\section{Related work}\label{sec2}
The current mainstream semantic segmentation models are divided into two main structures, CNN and Transformer methods. \\
    \textbf{CNN}: CNN mainly uses the convolutional layer to extract image feature information by sliding in the image to obtain the whole image feature information to classify the image. Semantic segmentation is a particular form of image classification used to classify the image at the pixel level. It is not difficult to understand why many network models use the model of image classification as the backbone. Use the mature image classification The use of established image classification algorithms to further manipulate the features extracted from images to achieve pixel-level image classification can achieve better performance and simplify network design. To meet the need for pixel-level features for semantic segmentation, PSPNet\upcite{zhao2017pyramid} uses the Spatial Pyramid Pooling (SPP) module behind the backbone to adapt to objects of different sizes while can access to multi-scale feature information. UNet\upcite{ronneberger2015u} achieves a good segmentation effect by fusing semantic feature information from the higher and lower layers through jump connections. Still, its ability to extract feature information at each layer limits its performance. To extract more feature information at each layer, ENet\upcite{paszke2016enet} fuses convolutional operations and pooling operations to obtain feature information during downsampling, CGNet\upcite{wu2020cgnet} uses Context Guided(CG) block to combine normal convolutional operations and null convolution to obtain contextual feature information and enhances feature extraction by adding an attention structure. BiSeNet\upcite{yu2018bisenet} uses a two-branch structure to fuse spatial and contextual feature information. The semantic segmentation structure in the form of CNN without backbone is more about better feature extraction at each layer. In contrast, the segmentation network structure with backbone is more about obtaining deeper feature information after the backbone using a multi-branch structure for better segmentation. \\  
    \textbf{Transformer}: Transformer originated from the natural language process(NLP). Because of its outstanding performance in NLP, researchers in computer vision paid attention to it. Soon, a series of related works\upcite{dosovitskiy2020image}\upcite{carion2020end}\upcite{zheng2021rethinking} applied Transformer to semantic segmentation, and these works got good segmentation results. In a special structural form, a transformer can cover every pixel value in the image and obtain global feature information, but it brings colossal computation. Swin Transformer\upcite{liu2021swin} proposed a solution to this problem, cutting the image into pieces and extracting features according to first local and then whole. It made good use of CNN's idea of locality, significantly reducing the computation, but compared with CNN, it still has a big gap, and its hardware environment requires high requirements, and reasoning speed is slow. To summarize, we propose a Patch block, which incorporates Swin Transformer's slicing idea into CNN structure in a different way to better solve the inadequacies of Transformer and CNN. For reference, we use CNN's module to extract context information so that our module may extract more extensive feature information at each layer of the network.

\begin{figure}[htb]
	\centering
	\includegraphics[width=\textwidth]{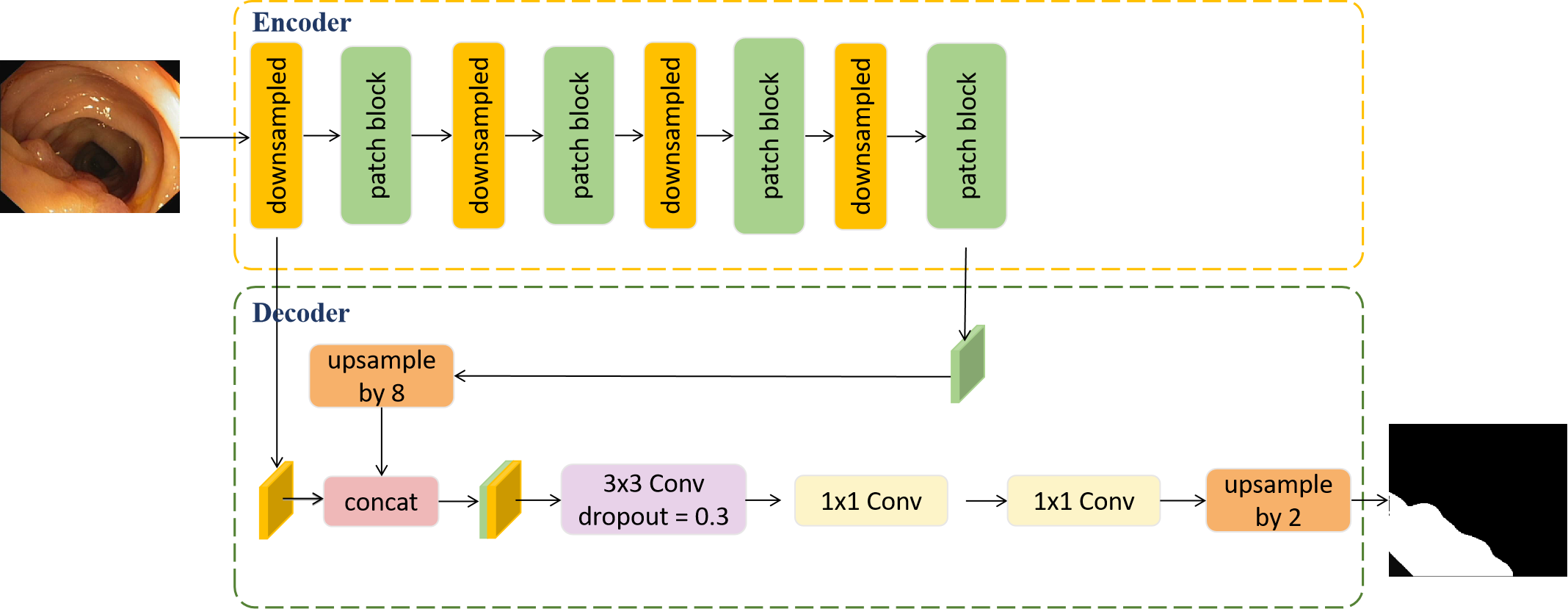}
	\caption{Proposed PNet architecture.}
	\label{fig.PNet}
\end{figure}

\section{Methodology}\label{sec3}
\sloppy
In this work, we propose Patch Network (PNet), which adopts the classical encoder-decoder structure. In this section, we describe the two parts of encoder and decoder, describe the components of PNet, and compare similar units of Patch block.
\subsection{Encoder}\label{subsec31}
\begin{figure*}[htb]
\centering
	\begin{minipage}[b]{0.35\linewidth}
		\centering
		\includegraphics[width=4cm]{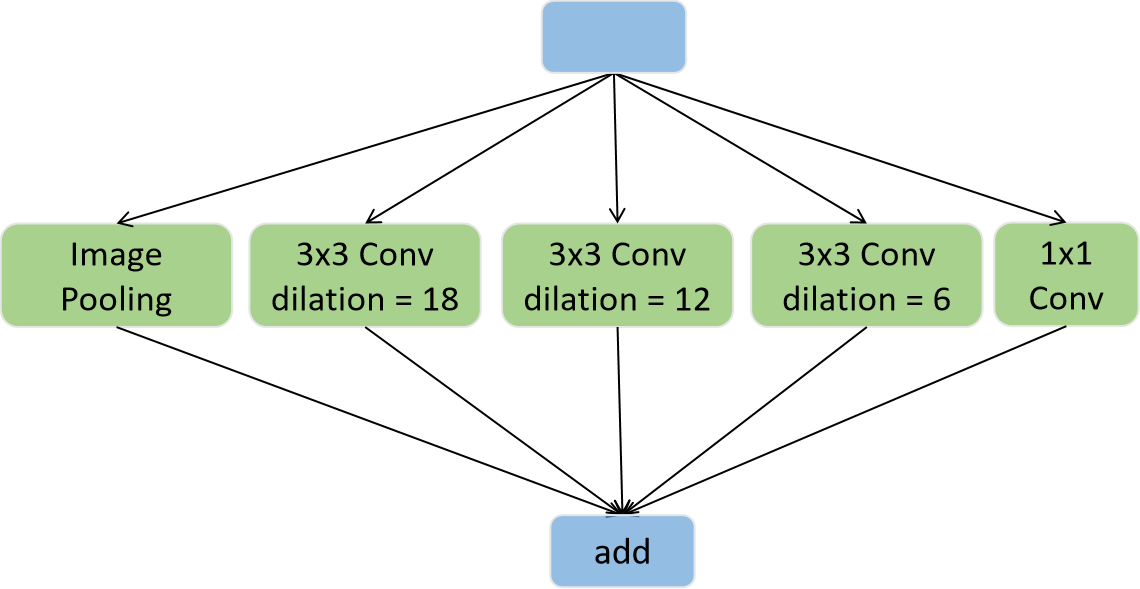}
		\centerline{(a) ASPP}
		\label{aspp}
	\end{minipage}
	\begin{minipage}[b]{0.3\linewidth}
		\centering
		\includegraphics[width=3cm]{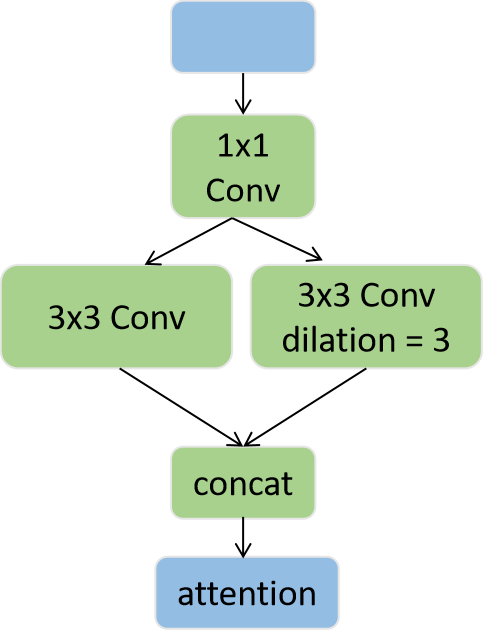}
		\centerline{(b) CG block}
		\label{CG block}
	\end{minipage}
	\begin{minipage}[b]{0.3\linewidth}
		\centering
		\includegraphics[width=3cm]{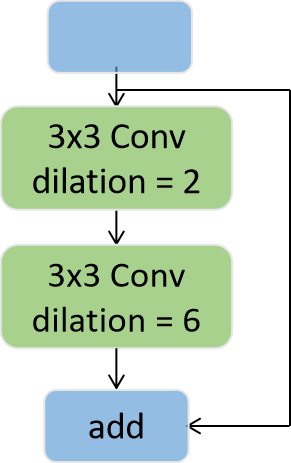}
		\centerline{(c) Patch block}
		\label{pa block}
	\end{minipage}
	\caption{Comparison of different feature extractors.}
	\label{fig:block}
\end{figure*}

It is mainly composed of a downsampling module and Patch block at this stage. Compared with max-pooling or average pooling, which directly retains features through simple extraction methods, we prefer to use convolution operations to achieve downsampling operations through learning. Many experiments have proved that the downsampling operation implemented by convolution is better than the pooling operation. At the same time, inspired by\upcite{ding2022scaling}, we did not use the conventional 3x3 convolution kernel. And the bold attempt to use a 5x5 convolution kernel got better performance, so the downsampling module is implemented by a 5x5 convolution kernel, the stride of 2, and padding of 2. For the CNN network model, the ability of each layer to extract image features dramatically affects the final performance of the model. For example, the classic network deeplab\upcite{chen2018encoder} that uses an image classification network as a backbone, the default image classification network, has learned good features. Next, using aspp in the decoder stage (as shown in Fig~\ref{fig:block}) further complements the contextual features through a larger receptive field, achieving good results. Like CGNet, which does not use a backbone network model, it is necessary to design a module to extract feature information. It uses a CG block (as shown in Fig~\ref{fig:block} ) dual-branch structure to enhance the feature extraction capability of each layer to improve the performance of the model. The original intention of our design was to use a module with aspp feature extraction capability to replace CG block, and the emergence of Swin Transformer has become the spark of our module design. Its idea of learning image blocks inspired us. Since the implementation and characteristics of CNN and Transformer are different, copying their structure is too poor to explain. We designed a module that first uses convolution with a small atrous rate to learn the features and then uses a large atrous. The convolution of the rate is used for further learning, which is similar to the form of "patching," so we call it a Patch block(as shown in Fig~\ref{fig:block}), and the two atrous rates are set to 2 and 6, respectively. For intuitive understanding, we show a standard 3x3 convolution (Fig~\ref{fig.atrous}), a 3x3 convolution with both atrous rate and padding of 2 (Fig~\ref{fig.atrous}), and a 3x3 convolution with both atrous rate and padding of 6 (Fig~\ref{fig.atrous}), We temporarily call the standard 3x3 convolution area the standard area. In the first atrous convolution, it can be seen that some areas of the standard area are used, and the surrounding areas are jointly learned, and the sliding window process will combine the upper three areas and the lower three areas. Learn each area, and then use the second atrous convolution with a more significant atrous rate to learn the middle area, and at the same time learn with a broader range of context information, and finally use the original input and the result here to perform residual learning through add, Further improve the feature extraction ability and make up for the feature information of the missing area.
\subsection{Decoder}\label{subsec32}
Considering the design of the lightweight model and the powerful feature extraction capability of the Encoder stage, we have simplified the design of the Decoder. First, perform an 8-fold upsampling operation on the features learned in the previous step and concat them with the first downsampled features to fuse the deep semantic information and shallow spatial information. Since Patch block also brings about the redundant information, so we first use 3x3 convolution to learn deep and shallow fusion information features and then use a dropout of 0.3 to suppress redundant information, which also prevents the model from overfitting to a certain extent and then use two 1x1 convolution, the former is optimized across channels, the latter outputs the number of classification categories, and finally returns to the original image input size using an upsampling operation again.
\subsection{Patch Network}\label{subsec33}
Four downsampling operations are performed in the Encoder stage. After each downsampling, the Patch block is used to obtain rich contextual information. In the decoder stage, the deep semantic information and shallow spatial information are fused, and dropout is used to prevent overfitting. At this point, a well-designed lightweight CNN network for medical image segmentation was born (As shown in Fig~\ref{fig.PNet}).

\begin{figure*}[htb]
\centering
	\begin{minipage}[b]{0.3\linewidth}
		\centering
		\includegraphics[width=3cm]{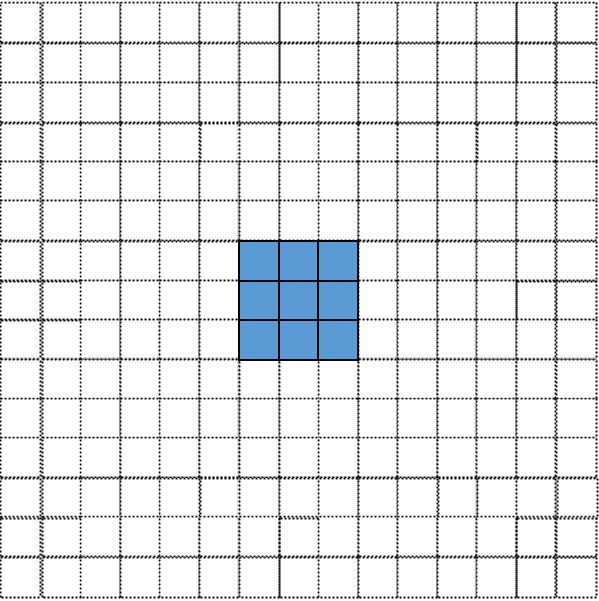}
		\centerline{(a) atrous rate=1(blue)}
		\label{rate1}
	\end{minipage}
	\begin{minipage}[b]{0.3\linewidth}
		\centering
		\includegraphics[width=3cm]{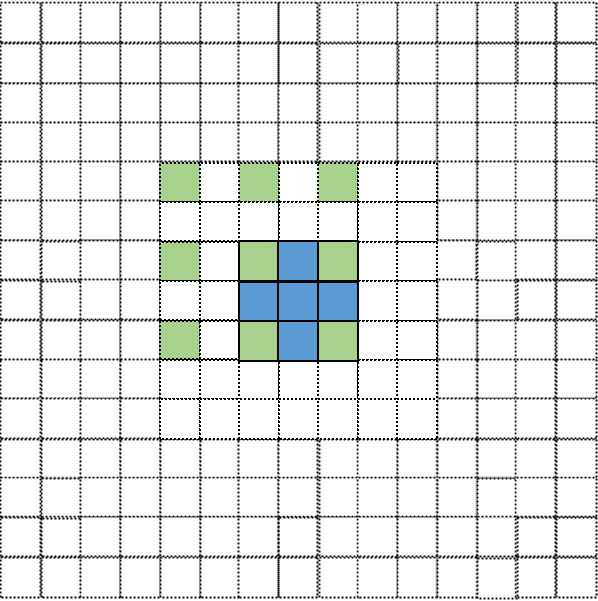}
		\centerline{(b) atrous rate=2(green)}
		\label{rate2}
	\end{minipage}
	\begin{minipage}[b]{0.3\linewidth}
		\centering
		\includegraphics[width=3cm]{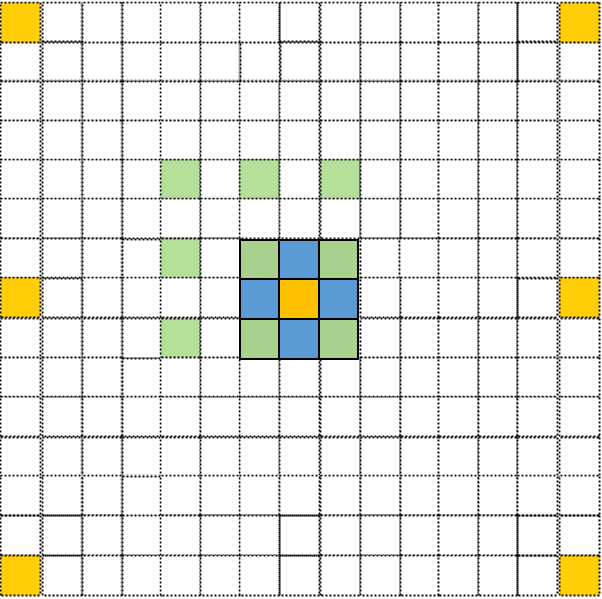}
		\centerline{(c) atrous rate=6(yellow)}
		\label{rate3}
	\end{minipage}
	\caption{atrous convolution.}
	\label{fig.atrous}
\end{figure*}

\section{Experiments and Results}\label{sec4}

\subsection{Datasets}\label{subsec41}

\subsubsection{Polyp Segmentation}\label{subsubsec42}
Accurate detection of colon polyps is of great significance for colon cancer prevention. CVC-ClinicDB[2](CVC for short)\upcite{Bernal2015WMDOVAMF} includes 612 colon polyp images. We use the original size $384 \times $288 image and split it into train set(80\%) and test set(20\%).

\subsubsection{ETIS}\label{subsubsec43}
Like CVC, ETIS - LaribPolypDB(ETIS for short) is also a polyp dataset containing 196 images from 29 sequences, all from different devices. We resize all the original images to $512 \times $384 and split them into train set(80\%) and test set(20\%).

\subsubsection{Skin Lesion Segmentation}\label{subsubsec44}
Computer-aided automatic diagnosis of Skin cancer is an inevitable trend, and Skin lesion segmentation is urgent as the first step. The data set is from MICCAI 2018 Workshop - ISIC2018: Skin Lesion Analysis Towards Melanoma Detection[8][14](Skin for short). It contains 2594 images and is randomly split into train set (80\%) and test set (20\%). For better model training and result display, we resize all the original images to $224 \times $224.

\subsection{Implementation details}\label{subsec45}
For three benchmarks and multiple segmentation models, we set consistent training parameters. We set epochs as 200 in the three data sets. We use a learning rate(LR) equal to 1e-4 for all tasks. In addition, we use batch size equal to 2 for ETIS and CVC tasks and 4 for the Skin task. Cross entropy loss and Adam are used as loss function and optimizer, respectively. All experiments run on the NVIDIA TITAN V GPU with 12GB. Intersection over Union (IOU), Dice coefficient, FPS, and computational complexity(FLOPs) are selected as the evaluation metrics in this paper. We used these evaluation metrics for all datasets. At the same time, we use the random rotation of 90 degrees, mirror surface, random brightness, random contrast, and other methods to perform data enhancement on these training data, increase the data set to prevent overfitting, and improve the robustness of the model.

\begin{table}[htbp]
    \caption{Evaluation of proposed PNet}
    \resizebox{\textwidth}{!}{
    \begin{tabular}{cccccccccccc}
    \hline
        Dataset & Methods & IOU & Dice & Params(M) & FLOPs(G) & FPS &   \\ \hline
        CVC & UNet & 0.8713 & 0.9174 & 34.53 & 110.5 & 21.83 &   \\  
          & PSPNet & 0.9152 & 0.9491 & 60.09 & 100.84 & 13.63 &   \\  
          & SegNet\upcite{badrinarayanan2017segnet} & 0.8790 & 0.9252 & 29.44 & 67.67 & 21.28 &   \\  
          & AttU\_Net\upcite{oktay2018attention} & 0.8458 & 0.9021 & 34.87 & 112.34 & 18.26 &   \\  
          & DenseUnet\upcite{cao2020denseunet} & 0.9209 & 0.9541 & 19.33 & 26.01 & 14.47 &   \\  
          & DoubleUNet\upcite{jha2020doubleu} & 0.9219 & 0.9533 & 18.84 & 74.52 & 16.02 &   \\  
          & UNet++\upcite{zhou2019unet++} & 0.9046 & 0.9424 & 36.63 & 232.98 & 13.53 &   \\  
          & PNet & \textbf{0.9332} & \textbf{0.9599} & \textbf{3.38} & \textbf{23.4} & \textbf{44.34} &   \\ \hline
        ETIS & UNet & 0.8708 & 0.9226 & 34.53 & 196.45 & 7.68 &   \\  
          & PSPNet & 0.9135 & 0.9508 & 60.09 & 179.23 & 7.52 &   \\  
          & SegNet & 0.6477 & 0.7361 & 29.44 & 120.31 & 9.09 &   \\  
          & AttU\_Net & 0.8335 & 0.8973 & 34.87 & 199.72 & 7.13 &   \\  
          & DenseUnet & 0.9274 & 0.9602 & 19.33 & 46.24 & 7.40 &   \\  
          & DoubleUNet & 0.9040 & 0.9442 & 18.84 & 132.47 & 7.45 &   \\  
          & UNet++ & 0.8809 & 0.9254 & 36.63 & 414.19 & 5.37 &   \\  
          & PNet & \textbf{0.9405} & \textbf{0.9646} & \textbf{3.38} & \textbf{41.61} & \textbf{16.39} &   \\ \hline
        Skin & UNet & 0.8681 & 0.9159 & 34.53 & 100.2 & 59.19 &   \\  
          & PSPNet & 0.8887 & 0.9300 & 60.09 & 91.5 & 43.55 &   \\  
          & SegNet & 0.8586 & 0.9101 & 29.44 & 30.7 & 85.32 &   \\  
          & AttU\_Net & 0.8458 & 0.9021 & 34.87 & 50.97 & 60.36 &   \\  
          & DenseUnet & 0.8844 & 0.9289 & 19.33 & 11.8 & 66.57 &   \\  
          & DoubleUNet & 0.8832 & 0.9256 & 18.84 & 33.81 & 56.85 &   \\  
          & UNet++ & 0.8837 & 0.9276 & 36.63 & 105.7 & 41.04 &   \\  
          & PNet & \textbf{0.8946} & \textbf{0.9340} & \textbf{3.38} & \textbf{10.62} & \textbf{126.23} &   \\  \hline
    \label{table:QC}
    \end{tabular}}
\end{table}

\subsection{Experimental Results}\label{subsec46}
\begin{figure*}[!htp]
	\begin{minipage}[b]{0.3\linewidth}
		\centering
		\includegraphics[width=1.2cm]{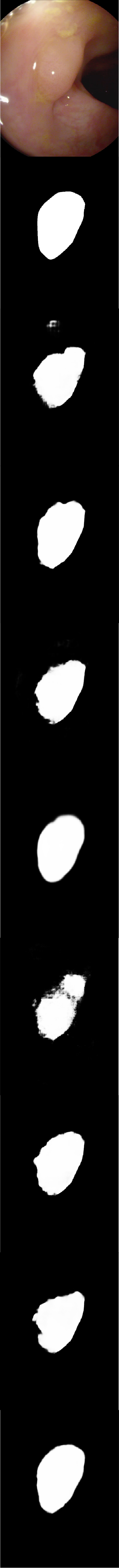}
		\centerline{(a) ETIS}
	\end{minipage}
	\begin{minipage}[b]{0.3\linewidth}
		\centering
		\includegraphics[width=1.2cm]{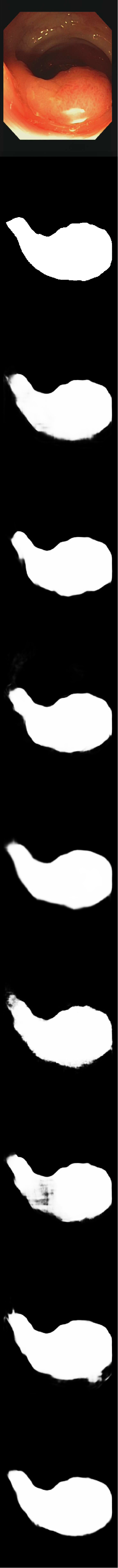}
		\centerline{(b) CVC}
	\end{minipage}
	\begin{minipage}[b]{0.3\linewidth}
		\centering
		\includegraphics[width=1.59cm]{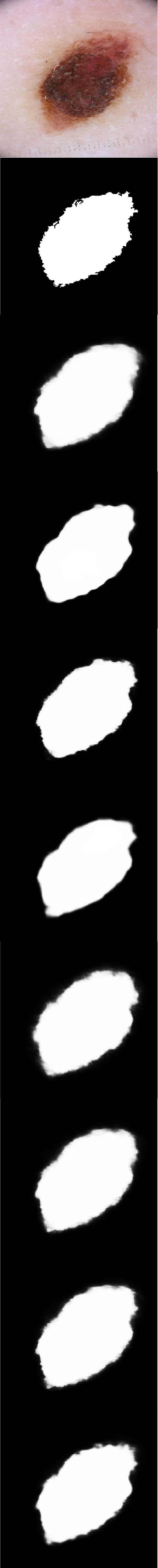}
		\centerline{(c) ISIC2018}
	\end{minipage}
	\caption{Qualitative comparison of segmentation results for CVC, ETIS, and Skin datasets, from top to bottom are Image, Ground Truth, U-Net, PSPNet, SegNet, AttU-Net, DenseUnet, DoublueUNet, U-Net++, PNet}
	\label{fig:QC}
\end{figure*}

To further demonstrate the superiority of our model, we evaluate it on three datasets and use the four evaluation metrics of IOU, Dice, FLOPs, and FPS to show the performance comparison of multiple models. The quantitative results are shown in Table~\ref{table:QC}. At the same time, we also show the results visualization in Fig~\ref{fig:QC}. results of polyp datasets As shown in Table 1, our model significantly outperforms other models on both polyp datasets, CVC and ETIS, especially on ETIS, a small dataset with only 196 images. Our model is superior to IOU and Dice They reached 0.9405 and 0.9646, respectively, which greatly surpassed other models, indicating that our model still has good performance in small datasets. In terms of model parameters and FLOPs, our model is also more lightweight. Compared with other models, it is only 1/10 of UNet++, and the FPS is three times faster than Unet++. This comparison is also reflected in the visualization results. From Fig~\ref{fig:QC}, we can see that the segmentation effect of our model is better than other models in the overall and edge performance, which is more evident in the ETIS dataset. The segmentation is rough, our model is the closest to the real mask, and the edge processing is perfect. results of Skin lesion Segmentation From the comparison results of the Skin dataset in Table 1, our model outperforms all other models on the IOU and Dice evaluation indicators. Our model parameters and FLOPs are much smaller than all other models, making our model has a faster FPS. The performance of the evaluation indicators of some models is close to our model, the best-performing model is PSPNet. Still, from the visualization results in Fig~\ref{fig:QC}, our model still has more obvious advantages. Other models, including PSPNet, segment the edges too much Smoothing, there is a big gap with the real mask, and this segmentation effect on complex edges extensively tests the segmentation ability of the model because the Patch block in our model can capture the feature extraction ability of more prominent context information, The segmentation performance is greatly improved, and the segmentation performance at the edge can still be close to the real mask.

\begin{table}[htbp]
    \caption{Ablation study on the shifted windows approach and different position embedding methods on three benchmarks, using
the PNet: 3x3 convlution, 3x3 convlution and maxpooling, 5x5 convlution}
    \resizebox{\textwidth}{!}{
    \begin{tabular}{cccccccccccc}
    \hline
        Dataset & \multicolumn{2}{c}{CVC} & \multicolumn{2}{c}{ETIS} & \multicolumn{2}{c}{Skin} & \\ \hline
        Methods & IOU & Dice & IOU & Dice & IOU & Dice & \\ 
        PNet(3X3) & 0.9176 & 0.9501 & 0.8938 & 0.9364 & 0.8820 & 0.9240 & \\ 
        PNet(pool) & 0.9283 & 0.9565 & 0.9258 & 0.9567 & 0.8809 & 0.9242 & \\
        PNet(5X5) & \textbf{0.9332} & \textbf{0.9599} & \textbf{0.9405} & \textbf{0.9646} & \textbf{0.8946} & \textbf{0.9340} & \\ \hline
    \label{table.conv}
    \end{tabular}}
\end{table}

\begin{table}[htbp]
    \caption{Ablation study on the shifted windows approach and different position embedding methods on three benchmarks, using Patch block to set different void rates for the first and second convolutions: 2 and 5, 2 and 6, 2 and 7, 3 and 8}
    \resizebox{\textwidth}{!}{
    \begin{tabular}{cccccccc}
        \hline
        Dataset & CVC &  & ETIS &  & Skin &  &   \\ \hline
        Methods & IOU & Dice & IOU & Dice & IOU & Dice  \\ 
        PNet(25) & 0.9187 & 0.9506 & 0.9308 & 0.9594 & 0.8653 & 0.9134 \\
        PNet(26) & \textbf{0.9332} & \textbf{0.9599} & \textbf{0.9405} & \textbf{0.9646} & \textbf{0.8946} & \textbf{0.9340} \\ 
        PNet(27) & 0.9289 & 0.9573 & 0.9287 & 0.9579 & 0.8885 & 0.9300 \\
        PNet(38) & 0.9166 & 0.9501 & 0.9048 & 0.9443 & 0.8898 & 0.9292 \\ \hline
    \end{tabular}}
    \label{table.atrous}
\end{table}

\subsection{Ablation Study }\label{subsec47}
To further confirm the effectiveness of our proposed module, we conduct ablation experiments on the proposed module. First, the atrous rate of the two atrous convolutions in our model's core module Patch block is verified. From the experimental results in Table~\ref{table.conv}, it can be obtained that the atrous rates of our proposed two atrous convolutions are 2 and 6, respectively. The best effect is consistent with the image interpretation in Fig~\ref{fig.atrous}. When the atrous rate of the first convolution is 2, we compare the performance of the second convolution with the atrous rate of 5, 6, and 7. The second dilation rate needs to cover the first convolution range according to the design concept. When the first dilation ratio is 2, the convolution range is 5x5, and when the dilation ratio is 6, that is, every two. The distance between the convolution kernels is 5, which covers the first convolution range. Similarly, when the first convolution rate is 3, the second convolution rate should be 8. The experiment was also carried out as a comparison, and the segmentation performance was not good. The reason for the guess is that the convolution range is too extensive. For some small resolution images, the image size after four times of downsampling will be tiny, and the convolution kernel is 3. When the dilation ratio is 8, the convolution range is 17x17. At this time, the convolution kernel may be larger than the image itself, resulting in the need to fill too many boundaries for calculation, which will result in poor results. In addition, we also conducted a comparative experiment on the downsampling module in the Encoder stage, using a 3x3 convolution kernel with a stride of 2, and using a 3x3 convolution combined with maximum pooling for comparison, as can be seen from the experimental results in Table~\ref{table.conv}, Our downsampling module achieves optimality, and these ablation experiments demonstrate the performance of our module.

\section{Conclusion}\label{sec5}
Conclusion In this paper, we propose an efficient feature extraction module Patch block, and based on it, and we propose a Patch Network for medical image segmentation. The IOU and Dice obtained by our experiments on three benchmarks are significantly better than other models. The segmentation performance on the image edge shows the superior feature extraction ability of the Patch block. The experimental results of ETIS also show that our model works on a small data set of outstanding performance. The model size and FLOPs are only 1/10 of UNet++, and the inference speed is still better than three times.
%
%


%
%

\bibliographystyle{spbasic}      
\bibliography{sn-bibliography}   


\end{document}